\documentclass{article}
\usepackage[T1]{fontenc}
\usepackage[utf8]{inputenc}
\usepackage{verbatim}
\usepackage{wrapfig}
\usepackage{booktabs}
\usepackage{graphicx}
\usepackage{microtype}

\usepackage{hyperref}
 
\hypersetup{unicode=true, bookmarks=false, breaklinks=false,pdfborder={0 0 0},backref=section,colorlinks=false, hidelinks=true}

\makeatletter

\providecommand{\tabularnewline}{\\}

\usepackage[final, nonatbib]{neurips_2020}

\usepackage[utf8]{inputenc} 
\usepackage[T1]{fontenc}    
\usepackage{hyperref}       
\usepackage{url}            
\usepackage{booktabs}       
\usepackage{amsfonts}       
\usepackage{nicefrac}       
\usepackage{microtype}      
\usepackage{graphicx}       
\usepackage{amsmath}        
\usepackage{multirow}

\title{Biomedical Information Extraction for\\Disease Gene Prioritization}

\author{
  Jupinder Parmar\thanks{Work conducted while author was an intern at OccamzRazor.} \ \thanks{Equal contribution.} \\
  Stanford University\\
  \texttt{jsparmar@stanford.edu}
  \And
  William Koehler\footnotemark[2]\\
  OccamzRazor\\
  \texttt{william@occamzrazor.com}
  \And
  Martin Bringmann\\
  OccamzRazor\\
  \texttt{martin@occamzrazor.com}
  \And
  Katharina Sophia Volz\\
  OccamzRazor\\
  \texttt{volz@occamzrazor.com}
  \And
  Berk Kapicioglu\footnotemark[2] \\
  OccamzRazor\\
  \texttt{berk@occamzrazor.com}
}

\makeatother

\begin{document}
\maketitle
\begin{abstract}
We introduce a biomedical information extraction (IE) pipeline that
extracts biological relationships from text and demonstrate that its
components, such as named entity recognition (NER) and relation extraction
(RE), outperform state-of-the-art in BioNLP. We apply it to tens of
millions of PubMed abstracts to extract protein-protein interactions
(PPIs) and augment these extractions to a biomedical knowledge graph
that already contains PPIs extracted from STRING, the leading structured
PPI database. We show that, despite already containing PPIs from an
established structured source, augmenting our own IE-based extractions
to the graph allows us to predict novel disease-gene associations
with a 20\% relative increase in hit@30, an important step towards
developing drug targets for uncured diseases.
\end{abstract}

\section{Introduction}

Understanding diseases and developing curative therapies requires
extracting and synthesizing relevant knowledge from vast swaths of
biomedical information. However, with the exponential growth of scientific
publications over the past several decades \cite{Landhuis2016}, it
has become increasingly difficult for researchers to keep up with
them. Moreover, most biomedical information is only disseminated via
unstructured text, which is not amenable to most computational methods
\cite{Huang2016}. Thus, there is a growing need for scalable methods
that can both extract relevant knowledge from unstructured text and
synthesize it to infer novel biomedical discoveries.%

To fill this need, we build an end-to-end biomedical IE pipeline \cite{Huang2016,Gonzalez2016,Gachloo2019}
by leveraging SciSpacy \cite{Neumann2019}, the most modern and actively
developed open-source BioNLP library, and customizing its NER and
RE components via transfer learning and BioBERT \cite{Lee2020,Devlin2019}.
We demonstrate that our pipeline %
{}outperforms the existing state-of-the-art (SOTA) %
{}for biomedical IE, such as PubTator Central \cite{Wei2019}, its
RE extensions \cite{Percha2018}, and SciSpacy \cite{Neumann2019}
itself. %

We then run our pipeline on the PubMed \cite{Canese2013} corpus,
the largest repository of biomedical abstracts, and extract protein-protein
interactions (PPI). Even though our pipeline can easily be trained
to extract any relationship, we focus on PPIs because our understanding
of them is only partially complete \cite{Stumpf2008,Venkatesan2009,Vidal2011},
they play an important role in identifying novel disease-gene associations
\cite{Moreau2012}, and there is already an established structured
PPI database called STRING \cite{Szklarczyk2019} that allows us to
benchmark our extractions. 

Finally, we augment our IE-based PPIs to a knowledge graph that already
contains STRING-based PPIs and demonstrate that the augmentation yields
a 20\% relative increase in hit@30 for predicting novel disease-gene
associations. Even though biomedical IE pipelines have previously
been evaluated in downstream link prediction tasks when the IE-based
extractions were the sole source of the graph \cite{Fauqueur2019,Nagarajan2015a},
to the best of our knowledge, we are the first to show a lift in a
setting where the knowledge graph is already populated by an established
structured database that contains the same relation type.

Increasing predictive accuracy in such a difficult setting demonstrates
the quality of our biomedical IE pipeline, which is specifically designed
to require only a small amount of training data to extract any biomedical
relationship, and moves us one step closer towards developing drug
targets for uncured diseases.%

\section{Biomedical Information Extraction}

\begin{figure}
\centering{}\includegraphics[scale=0.35]{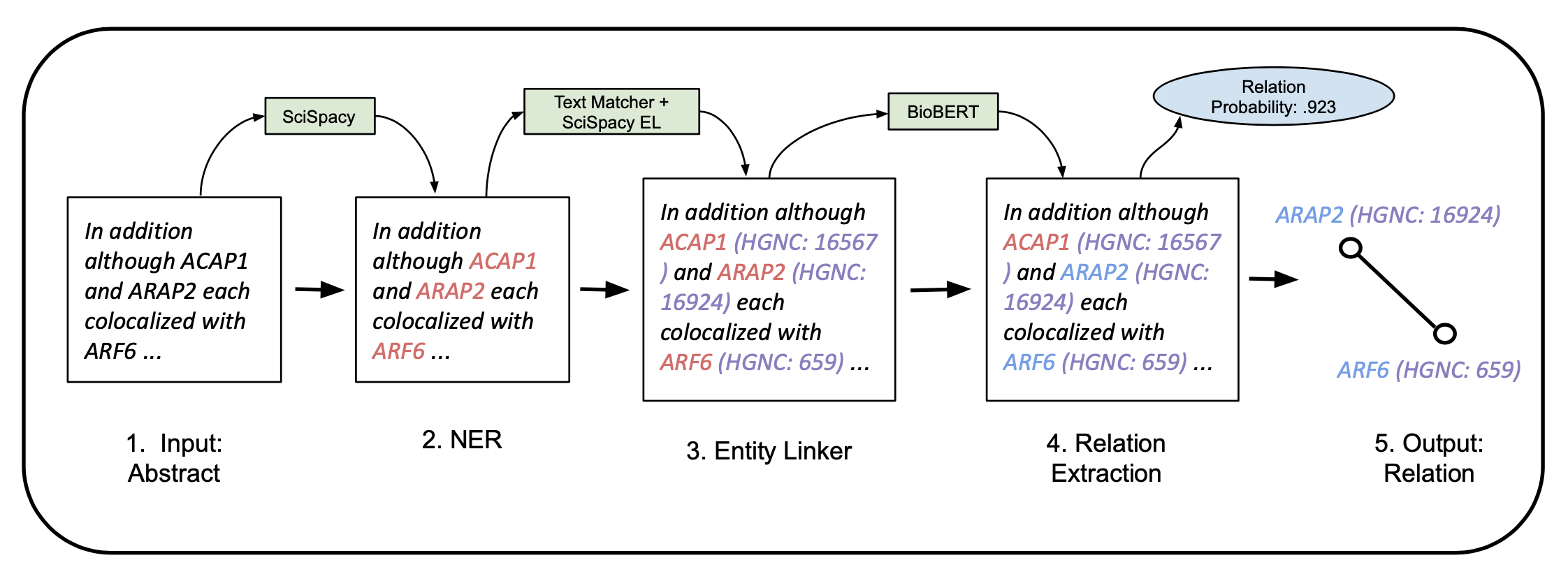}{\footnotesize{}\caption{{\footnotesize{}\label{fig:ie-pipeline}A high-level overview of our
IE pipeline. We only display the single candidate relation (ARAP2,
ARF6) for simplicity although three candidate relations are present.}}
}
\end{figure}

In Figure \ref{fig:ie-pipeline}, we provide an overview of our biomedical
IE pipeline that we train and evaluate on PPI data annotated by in-house
biologists. In the following subsections we review how we configured
the pipeline for biomedical text and show how each component outperforms
its leading competitor in BioNLP. 

\subsection{Named Entity Recognition (NER)}

\begin{wraptable}{r}{0pt}%
\centering{}%
\begin{tabular}{lccc}
\toprule 
System & Precision & Recall & F1\tabularnewline
\midrule 
Our Model & \textbf{78.41} & \textbf{73.87} & \textbf{76.08}\tabularnewline
PubTator & 58.96 & 49.20 & 45.76\tabularnewline
ScispaCy & 37.81 & 57.96 & 53.64\tabularnewline
\bottomrule
\end{tabular}\caption{{\footnotesize{}\label{tab:ner-results} NER Test Results}}
\end{wraptable}%

We train our NER model using SpaCy \cite{Honnibal2017}, which we
customize further via ScispaCy's \cite{Neumann2019} word vectors
pre-trained on biomedical text. Our training dataset consists of \textasciitilde 2000
PubMed abstracts tagged with proteins. We enforce strict annotation
rules during the labeling process to help disambiguate unclear protein
references, a task that we found other NER datasets do not accomplish
effectively given the complex nature of biomedical literature. We
then compare our model's performance on the test set against two of
the leading biomedical NER systems: PubTator Central \cite{Wei2019},
a web service that performs NER on PubMed abstracts, and ScispaCy
\cite{Neumann2019}, which provides its own protein NER model. As
seen in Table \ref{tab:ner-results}, our model outperforms both of
them.

\subsection{Relation Extraction (RE)}

For training and evaluating our RE model, we automatically annotate a separate set of
\textasciitilde 2000 PubMed abstracts using our NER model, generate
relation candidates between pairs of tagged proteins, and manually
annotate whether a given candidate contains an interaction. %
{}Using our NER model for annotation ensures that our RE model is trained
and evaluated based on the same data distribution it handles in production.

We then develop and evaluate a variety of RE models. First, we create
models based on feature engineering that use GloVe embeddings \cite{Pennington2014a}
and various linguistic features known to perform well on BioNLP tasks
\cite{Lever2017}. Then, we develop models based on BERT \cite{Devlin2019},
BioBERT \cite{Lee2020}, and SciBert \cite{Beltagy2020}. We represent
the task of relation extraction in these models using the entity start,
mention pool, and masked input configurations discussed in \cite{Lee2020,Soares2020}.
{}For BERT-based models, we experiment both with fine-tuning and feature
extraction. In our feature extraction experiments we combine BERT-based
features with our own engineered features.

\begin{wraptable}{l}{0pt}%
\centering{}%
\begin{tabular}{lccc}
\toprule 
System & Precision & Recall & F1\tabularnewline
\midrule 
v1 & \textbf{43.24} & 45.71 & 44.44\tabularnewline
v2 & 41.17 & 50.00 & \textbf{45.16}\tabularnewline
v3 & 31.37 & 68.57 & 43.04\tabularnewline
Masked BioBERT & 29.87 & \textbf{70.00} & 41.88\tabularnewline
\bottomrule
\end{tabular}\caption{{\footnotesize{}\label{tab:re-results}RE Test Results.}}
\end{wraptable}%

We compare each of our proposed configurations against the SOTA for
biological RE \cite{Lee2020}, a masked input BioBERT model. %
{}We refer to our top three models as v1: BioBert feature extraction
and feature engineering, v2: Fine-tuned SciBERT using mention pooling,
and v3: Fine-tuned BioBERT using entity start. Table \ref{tab:re-results}
reports the evaluation results for our top three models and the SOTA
model. We note that each of our models outperforms the SOTA model
in terms of the F1 score. Since all of the models perform well on
a different metric, we decide to run each of them on the entire PubMed
corpus.

\section{Extracting Relations from PubMed}

We run each of our pipeline configurations on PubMed \cite{Canese2013},
a repository of over 30 million biomedical abstracts that we filter
down to 10 million based on their relevance to humans or mice.

\begin{figure}[htp]
\centering{}\includegraphics[viewport=0bp 50bp 1391bp 475bp,clip,scale=0.28]{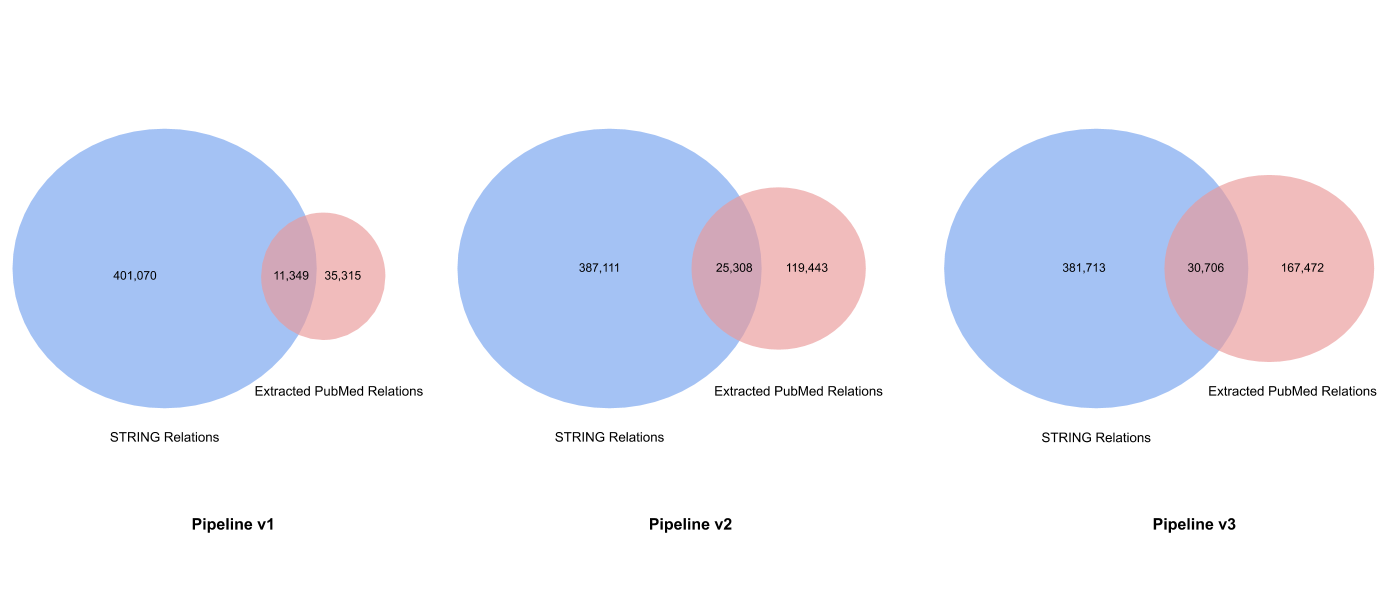}\caption{{\footnotesize{}\label{fig:pubmed-comparison}A comparison of different
IE pipelines and STRING.}}
\end{figure}

After extracting PPIs from PubMed, we compare them to the ones in
STRING \cite{Szklarczyk2019}, the leading structured PPI database,
and ascertain to what extent our IE-based extractions are novel and
in fact a segment of the siloed biomedical knowledge contained only
in text. The results of the comparison are shown in Figure \ref{fig:pubmed-comparison}.
We observe that IE-based PPIs do not significantly overlap with those
in STRING as the highest proportion of extracted relations contained in STRING among the three pipelines is v1 at 24.32\%. Additionally, we observe that each configuration behaves as we expect.
Specifically, pipeline v3, whose relation extraction model has the
highest recall, extracts the most relationships, whereas pipeline
v1, whose relation extraction model has the highest precision, extracts
the least relationships. 

Finally, our pipeline extracts more PPIs than previous biomedical
information extraction attempts. %
{}Most notably, Percha and Altman\cite{Percha2018} extend PubTator
\cite{Wei2019} with RE functionality by using a dependency parser
and clustering-based algorithms. They extract 41,418 PPIs, whereas
each of our pipelines extract substantially more. In addition, we
observe that the 198,178 PPIs pipeline v3 extracts is more in line
with the biomedical expectation since researchers determined there
to be roughly 650k PPIs in the human body of which only around 200k
have been validated experimentally \cite{Stumpf2008,Yeger-Lotem2015}.

\section{Disease Gene Prioritization}

The reason we developed our biomedical information extraction pipeline
is to extract biomedical knowledge from unstructured text, construct
a biomedical knowledge graph, and leverage this graph to infer novel
biomedical discoveries. In previous sections we demonstrated that
the components of our biomedical IE pipeline outperforms the leading
NER and RE models in BioNLP. In this section, we demonstrate that
our biomedical IE pipeline goes further and also enables novel biomedical
discoveries. 

Specifically, we focus on the problem of identifying disease genes,
 a set of genes associated with a particular disease. We formulate this
task as a link prediction problem \cite{Liben-Nowell2007,Martinez2016}
where we construct a biomedical knowledge graph and leverage the information
in the graph to predict previously unknown links between genes and
diseases. Identifying said links then helps in developing drug targets
for uncured diseases.

Historically, biomedical IE pipelines have been evaluated in downstream
link prediction tasks when the IE-based extractions were the sole
source of the graph \cite{Fauqueur2019,Nagarajan2015a}. In this paper,
we attempt to ascertain whether a biomedical IE pipeline can also
be used to complement an established structured database that provides
edges of the same relation type. 

To demonstrate this, we construct five different biomedical knowledge
graphs. For evaluation, we use DisGeNET \cite{Pinero2020}, the leading
database for gene-disease associations. We split DisGeNET edges randomly
into train (80\%), valid (10\%), and test sets (10\%), and use the
same valid and test sets for evaluating all five graphs. The only
difference between the five graphs is the training data. The first
graph only uses the train set of DisGeNET. The second graph augments
the train set of DisGeNET with STRING. The remaining graphs augment
the second graph, namely DisGeNET and STRING, with extractions from
one of the three versions of our biomedical IE pipeline. 

For each experiment, we train and evaluate a link prediction model
using a graph embedding algorithm called RotatE \cite{Sun2019} and
use a library called Optuna \cite{Akiba2019} for hyper-parameter
optimization. The results of the experiments are shown in Table \ref{tab:link-prediction-results}.
Note that MR is the mean of all gene-disease link ranks, MP is the
mean of the rank divided by the pool for that disease, and hit@k describes
the percentage of links we obtain in the top "k" ranks.

\begin{table}[ht]
\centering{}%
\begin{tabular}{lccccc}
\toprule 
 & MR & MP & hit@30 & hit@3 & hit@1\tabularnewline
\midrule 
IE v3 + STRING + DisGeNET & \textbf{1418.397} & \textbf{92.484} & \textbf{37.367}\% & \textbf{15.302}\% & \textbf{7.829}\%\tabularnewline
IE v2 + STRING + DisGeNET & 1441.802 & 92.262 & 35.409\% & 14.057\% & 7.473\%\tabularnewline
IE v1 + STRING + DisGeNET & 1829.548 & 89.869 & 32.74\% & 13.701\% & 6.762\%\tabularnewline
STRING + DisGeNET & 1952.084 & 89.362 & 31.139\% & 13.879\% & 7.651\%\tabularnewline
DisGeNET & 7422.117 & 59.544 & 0.356\% & 0.178\% & 0.178\%\tabularnewline
\bottomrule
\end{tabular}\caption{\label{tab:link-prediction-results} {\footnotesize{}Link prediction
results on various biomedical knowledge graphs.}}
\end{table}

We observe that augmenting v3 of our IE extractions to the graph provided
a lift across all metrics compared to the strong baseline of both
STRING and DisGeNET. Specifically, MR had a relative reduction of
27.3\%, hit@3 had a relative lift of 10.3\%, and hit@30 had a relative
lift of 20.0\%.

This indicates that the large amount of relations extracted from PubMed
contains high-quality edges and can be immediately helpful to a number
of biomedical tasks. Additionally, by achieving better performance
in disease gene identification when augmenting a knowledge graph that
already contained PPIs from a structured resource with our extracted
relations, we illustrate the tremendous representational power contained
in our IE-based PPIs. 

\section{Conclusion}

We have introduced a biomedical IE pipeline that can be configured
to extract any biomedical relationship from unstructured text using
a small amount of training data. We empirically demonstrated that
its NER and RE components outperform their leading competitors such
as PubTator Central \cite{Wei2019}, its RE extension \cite{Percha2018},
scispaCy \cite{Neumann2019}, and BioBERT \cite{Lee2020}. We then
ran it on tens of millions of PubMed abstracts to extract hundreds
of thousands of PPIs and show that these relations are novel in comparison
to the ones in leading structured databases. Finally, we evaluated
our IE-based PPIs' ability to enable biomedical discoveries by augmenting
them to a knowledge graph that already contains STRING-based PPIs
and showed that the augmentation yielded a 20\% relative increase
in hit@30 for predicting novel disease-gene associations. We believe that increasing
predictive accuracy in such a difficult setting demonstrates the quality
of our biomedical IE pipeline, which we plan to use to uncover other
biological relationships currently locked away in biomedical texts,
and moves us one step closer to developing drug targets for uncured
diseases.


\small{} \bibliographystyle{unsrt}
\bibliography{main}

\begin{thebibliography}{10}

\bibitem{Landhuis2016}
Esther Landhuis.
\newblock {Scientific literature: Information overload}.
\newblock {\em Nature}, 2016.

\bibitem{Huang2016}
Chung~Chi Huang and Zhiyong Lu.
\newblock {Community challenges in biomedical text mining over 10 years:
  Success, failure and the future}.
\newblock {\em Briefings in Bioinformatics}, 2016.

\bibitem{Gonzalez2016}
Graciela~H. Gonzalez, Tasnia Tahsin, Britton~C. Goodale, Anna~C. Greene, and
  Casey~S. Greene.
\newblock {Recent advances and emerging applications in text and data mining
  for biomedical discovery}.
\newblock {\em Briefings in Bioinformatics}, 2016.

\bibitem{Gachloo2019}
Mina Gachloo, Yuxing Wang, and Jingbo Xia.
\newblock {A review of drug knowledge discovery using BioNLP and tensor or
  matrix decomposition}.
\newblock {\em Genomics and Informatics}, 2019.

\bibitem{Neumann2019}
Mark Neumann, Daniel King, Iz~Beltagy, and Waleed Ammar.
\newblock {ScispaCy: Fast and Robust Models for Biomedical Natural Language
  Processing}.
\newblock In {\em BioNLP Workshop and Shared Task}, 2019.

\bibitem{Lee2020}
Jinhyuk Lee, Wonjin Yoon, Sungdong Kim, Donghyeon Kim, Sunkyu Kim, Chan~Ho So,
  and Jaewoo Kang.
\newblock {BioBERT: A pre-trained biomedical language representation model for
  biomedical text mining}.
\newblock {\em Bioinformatics}, 2020.

\bibitem{Devlin2019}
Jacob Devlin, Ming~Wei Chang, Kenton Lee, and Kristina Toutanova.
\newblock {BERT: Pre-training of deep bidirectional transformers for language
  understanding}.
\newblock In {\em North American Chapter of the Association for Computational
  Linguistics (NAACL)}, 2019.

\bibitem{Wei2019}
Chih~Hsuan Wei, Alexis Allot, Robert Leaman, and Zhiyong Lu.
\newblock {PubTator central: automated concept annotation for biomedical full
  text articles}.
\newblock {\em Nucleic Acids Research}, 2019.

\bibitem{Percha2018}
Bethany Percha and Russ~B. Altman.
\newblock {A global network of biomedical relationships derived from text}.
\newblock {\em Bioinformatics}, 2018.

\bibitem{Canese2013}
Kathi Canese and Sarah Weis.
\newblock {PubMed: The bibliographic database}.
\newblock {\em The NCBI Handbook}, 2013.

\bibitem{Stumpf2008}
Michael~P.H. Stumpf, Thomas Thorne, Eric {De Silva}, Ronald Stewart, Jun~An
  Hyeong, Michael Lappe, and Carsten Wiuf.
\newblock {Estimating the size of the human interactome}.
\newblock {\em Proceedings of the National Academy of Sciences (PNAS)}, 2008.

\bibitem{Venkatesan2009}
Kavitha Venkatesan, Jean~Fran{\c{c}}ois Rual, Alexei Vazquez, Ulrich Stelzl,
  Irma Lemmens, Tomoko Hirozane-Kishikawa, Tong Hao, Martina Zenkner, Xiaofeng
  Xin, Kwang~Il Goh, Muhammed~A. Yildirim, Nicolas Simonis, Kathrin Heinzmann,
  Fana Gebreab, Julie~M. Sahalie, Sebiha Cevik, Christophe Simon, Anne~Sophie
  de~Smet, Elizabeth Dann, Alex Smolyar, Arunachalam Vinayagam, Haiyuan Yu,
  David Szeto, Heather Borick, Am{\'{e}}lie Dricot, Niels Klitgord, Ryan~R.
  Murray, Chenwei Lin, Maciej Lalowski, Jan Timm, Kirstin Rau, Charles Boone,
  Pascal Braun, Michael~E. Cusick, Frederick~P. Roth, David~E. Hill, Jan
  Tavernier, Erich~E. Wanker, Albert~L{\'{a}}szl{\'{o}} Barab{\'{a}}si, and
  Marc Vidal.
\newblock {An empirical framework for binary interactome mapping}.
\newblock {\em Nature Methods}, 2009.

\bibitem{Vidal2011}
Marc Vidal, Michael~E. Cusick, and Albert~L{\'{a}}szl{\'{o}} Barab{\'{a}}si.
\newblock {Interactome networks and human disease}.
\newblock {\em Cell}, 2011.

\bibitem{Moreau2012}
Yves Moreau and L{\'{e}}on~Charles Tranchevent.
\newblock {Computational tools for prioritizing candidate genes: Boosting
  disease gene discovery}.
\newblock {\em Nature Reviews Genetics}, 2012.

\bibitem{Szklarczyk2019}
Damian Szklarczyk, Annika~L. Gable, David Lyon, Alexander Junge, Stefan Wyder,
  Jaime Huerta-Cepas, Milan Simonovic, Nadezhda~T. Doncheva, John~H. Morris,
  Peer Bork, Lars~J. Jensen, and Christian {Von Mering}.
\newblock {STRING v11: Protein-protein association networks with increased
  coverage, supporting functional discovery in genome-wide experimental
  datasets}.
\newblock {\em Nucleic Acids Research}, 2019.

\bibitem{Fauqueur2019}
Julien Fauqueur, Ashok Thillaisundaram, and Theodosia Togia.
\newblock {Constructing large scale biomedical knowledge bases from scratch
  with rapid annotation of interpretable patterns}.
\newblock In {\em BioNLP Workshop and Shared Task}, 2019.

\bibitem{Nagarajan2015a}
Meena Nagarajan, Angela~D. Wilkins, Benjamin~J. Bachman, Ilya~B. Novikov,
  Shenghua Bao, Peter~J. Haas, Mar{\'{i}}a~E. Terr{\'{o}}n-D{\'{i}}az, Sumit
  Bhatia, Anbu~K. Adikesavan, Jacques~J. Labrie, Sam Regenbogen, Christie~M.
  Buchovecky, Curtis~R. Pickering, Linda Kato, Andreas~M. Lisewski, Ana
  Lelescu, Houyin Zhang, Stephen Boyer, Griff Weber, Ying Chen, Lawrence
  Donehower, Scott Spangler, and Olivier Lichtarge.
\newblock {Predicting future scientific discoveries based on a networked
  analysis of the past literature}.
\newblock In {\em International Conference on Knowledge Discovery and Data
  Mining (KDD)}, 2015.

\bibitem{Honnibal2017}
Mattew Honnibal and Ines Montani.
\newblock {spaCy2: Natural language understanding with bloom embeddings,
  convolutional neural networks and incremental parsing.}
\newblock {\em To appear}, 2017.

\bibitem{Pennington2014a}
Jeffrey Pennington, Richard Socher, and Christopher~D. Manning.
\newblock {GloVe: Global vectors for word representation}.
\newblock In {\em Empirical Methods in Natural Language Processing (EMNLP)},
  2014.

\bibitem{Lever2017}
Jake Lever and Steven Jones.
\newblock {Painless Relation Extraction with Kindred}.
\newblock In {\em BioNLP Workshop and Shared Task}, 2017.

\bibitem{Beltagy2020}
Iz~Beltagy, Kyle Lo, and Arman Cohan.
\newblock {SCIBERT: A pretrained language model for scientific text}.
\newblock In {\em Empirical Methods in Natural Language Processing and
  International Joint Conference on Natural Language Processing
  (EMNLP-IJCNLP)}, 2020.

\bibitem{Soares2020}
Livio~Baldini Soares, Nicholas FitzGerald, Jeffrey Ling, and Tom Kwiatkowski.
\newblock {Matching the blanks: Distributional similarity for relation
  learning}.
\newblock In {\em Association for Computational Linguistics (ACL)}, 2020.

\bibitem{Yeger-Lotem2015}
Esti Yeger-Lotem and Roded Sharan.
\newblock {Human protein interaction networks across tissues and diseases}.
\newblock {\em Frontiers in Genetics}, 2015.

\bibitem{Liben-Nowell2007}
David Liben-Nowell and Jon Kleinberg.
\newblock {The link-prediction problem for social networks}.
\newblock {\em Journal of the American Society for Information Science and
  Technology}, 2007.

\bibitem{Martinez2016}
V{\'{i}}ctor Mart{\'{i}}nez, Fernando Berzal, and Juan~Carlos Cubero.
\newblock {A survey of link prediction in complex networks}.
\newblock {\em ACM Computing Surveys}, 2016.

\bibitem{Pinero2020}
Janet Pi{\~{n}}ero, Juan~Manuel Ram{\'{i}}rez-Anguita, Josep
  Sa{\"{u}}ch-Pitarch, Francesco Ronzano, Emilio Centeno, Ferran Sanz, and
  Laura~I. Furlong.
\newblock {The DisGeNET knowledge platform for disease genomics: 2019 update}.
\newblock {\em Nucleic Acids Research}, 2020.

\bibitem{Sun2019}
Zhiqing Sun, Zhi~Hong Deng, Jian~Yun Nie, and Jian Tang.
\newblock {RotatE: Knowledge graph embedding by relational rotation in complex
  space}.
\newblock In {\em International Conference on Learning Representations,
  (ICLR)}, 2019.

\bibitem{Akiba2019}
Takuya Akiba, Shotaro Sano, Toshihiko Yanase, Takeru Ohta, and Masanori Koyama.
\newblock {Optuna: A Next-generation Hyperparameter Optimization Framework}.
\newblock In {\em International Conference on Knowledge Discovery and Data
  Mining (KDD)}, 2019.

\end{thebibliography}
{\small\par}
\end{document}